\documentclass{article}

     \usepackage[preprint]{neurips_2019}

\usepackage[utf8]{inputenc} 
\usepackage[T1]{fontenc}    
\usepackage{hyperref}       
\usepackage{url}            
\usepackage{booktabs}       
\usepackage{amsfonts}       
\usepackage{nicefrac}       
\usepackage{microtype}      

\usepackage{times}
\usepackage{soul}
\usepackage{url}
\usepackage[utf8]{inputenc}
\usepackage[small]{caption}
\usepackage{graphicx}
\usepackage{amsmath}
\usepackage{booktabs}
\usepackage[ruled, vlined, linesnumbered]{algorithm2e}
\usepackage{microtype}
\usepackage{graphicx}
\usepackage{booktabs} 
\usepackage{hyperref}       
	\usepackage{amsfonts}       
\usepackage{nicefrac}       
\usepackage{blkarray}
\usepackage{float}
\usepackage{tikz}
\usetikzlibrary{calc}
\usepackage{amsthm}
\usepackage{dsfont}
\usepackage{mathrsfs}
\usepackage{amsmath}
\usepackage{autobreak}

\onecolumn

\usepackage{mathtools}
\usepackage{graphicx}
\graphicspath{ {figures/} }
\usepackage{subfig}

\newcommand{\bx}{\mathbf{x}}

\usepackage{enumitem}
\usepackage{diagbox}
\usepackage{bbm}
\usepackage{booktabs}

\title{Inductive Logic Programming via Differentiable Deep Neural Logic Networks}

\author{%
  Ali Payani\\
  Electrical and Computer Engineering\\
  Georgia Institute of Technology\\
  \texttt{payani@ece.gatech.edu} \\
  \And
  Faramarz Fekri \\
  Electrical and Computer Engineering\\
  Georgia Institute of Technology\\
  \texttt{Fekri@ece.gatech.edu} \\
}

\begin{document}

\maketitle

\begin{abstract}
  We propose a novel paradigm for solving Inductive Logic Programming (ILP) problems via deep recurrent neural networks.
This proposed ILP solver is designed based on differentiable implementation of the deduction via forward chaining. In contrast to the majority of past methods, instead of searching through the space of possible first-order logic rules by using some restrictive rule templates, we directly learn the symbolic logical predicate rules by introducing a novel differentiable Neural Logic (dNL) network. The proposed dNL network is able to learn and represent Boolean functions efficiently and in an explicit manner. We show that the proposed dNL-ILP solver supports desirable features such as recursion and predicate invention. Further, we investigate the performance of the proposed ILP solver in classification tasks involving benchmark relational datasets. In particular, we show that our proposed method outperforms the state of the art ILP solvers in classification tasks for Mutagenesis, Cora and IMDB datasets.

\end{abstract}

\section{Introduction}
\label{sec:introduction}
Despite the tremendous success of the deep neural networks, they are still prone to some limitations. These systems, in general, do not construct any explicit and symbolic representation of the algorithm they learn. In particular, the learned algorithm is implicitly stored in thousands or even millions of weights, which is typically impossible for human agents to decipher or verify. 
Further, MLP networks are suitable when large training examples are available. Otherwise, they usually do not generalize well.
One of the machine learning approaches that addresses these shortcomings is Inductive Logic Programming (ILP). In ILP, explicit rules and symbolic logical representations can be learned using only a few training examples. Further, the solutions usually generalize well. 
%

The idea of using neural networks for learning ILP has attracted a lot of research in recent years ( \cite{holldobler1999approximating,francca2014fast,serafini2016logic,evans2018learning}). Most neural ILP solvers work by propositionalization of the relational data and use the neural networks for the inference tasks. As such, they usually are superior to classical ILP solvers in handling missing or uncertain data. However, in many of the proposed neural solvers, the learning is not explicit (e.g. connectionist  network (\cite{bader2008connectionist}). Further, these methods do not usually support features such as inventing new predicates and learning recursive rules for predicates. Additionally, in almost all of the past ILP solvers, the space of possible symbolic rules for each predicate is significantly restricted and reduced  by introducing some types of rule templates before searching through this space for possible candidate. (e.g., mode declarations in Progol (\cite{muggleton1995inverse}) and meta-rules in Metagol). 
In fact, as stated in \cite{evans2018learning}, the need for using program templates to generate a limited set of viable candidate clauses in forming the predicates is the key weakness in all existing (past) ILP systems (neural or non-neural), severely limiting the solution space of a problem. 
The contribution of this paper is as follows: we introduce a new neural framework for learning ILP, by using a differentiable implementation of the forward chaining. Further, we practically remove the need for the use of rule templates by introducing novel symbolic Boolean function learners via multiplicative neurons. This flexibility in learning the first-order formulas without the need for a rule template makes it possible to learn very complex recursive predicates. Finally, as will show in the experiments, the proposed method outperforms the state of the art ILP solvers in relational data classification for the problems involving thousands of constants.

\section{Inductive Logic Programming via dNL}
\label{sec:ILP}
Logic programming is a programming paradigm in which we use formal logic (and usually first-order-logic) to describe relations between facts and rules of a program domain. 
In this framework rules are usually written as clauses of the form:
\begin{equation}\label{eq:clause}
H \leftarrow B_1,\,B_2,\,\dots,\,B_m
\end{equation}
where $H$ is called \texttt{head} of the clause and $B_1,\,B_2,\,\dots,\,B_m$ is called \texttt{body} of the clause. A clause of this form expresses that if all the atoms in the \texttt{body} are true, the \texttt{head} is necessarily true.
We assume each of the terms $H$ and $B$ are made of \texttt{atoms}. Each \texttt{atom} is created by applying an $n$-ary Boolean function called \texttt{predicate} to some constants or variables. A \texttt{predicate} states the relation between some variables or constants in the logic program. Throughout this paper we will use small letters for constants and capital letters (A, B, C, ...) for variables. 
%
In ILP, a problem can be defined as a tuple ($\mathcal{B}$,$\mathcal{P}$,$\mathcal{N}$) where $\mathcal{B}$ is the set of background assumptions and $\mathcal{P}$ and $\mathcal{N}$ are the set of positive and negative examples, respectively. Given this setting, the goal of the ILP is to construct a logic program (usually expressed as a set of definite clauses, $\mathcal{R}$) such that it explains all the examples. More precisely, 
\begin{equation}
    \mathcal{B},\mathcal{R}\models e,\forall e \in \mathcal{P}\quad,\quad \mathcal{B},\mathcal{R}\not\models e,\forall e \in \mathcal{N}
\end{equation} 
Let's consider the logic program that defines the \texttt{lessThan} predicate over natural numbers and assume that our constants contains the set $\mathcal{C} = \{0,1,2,3,4\}$ and the ordering of the natural numbers are defined using the predicate \texttt{inc} (which defines increments of 1). The set of background atoms which describes the known facts about this problem is the set $\mathcal{B} = \{\text{inc}(0,1),\, \text{inc}(1,2),\, \text{inc}(2,3),\, \text{inc}(3,4)\}$. Further, $\mathcal{P}=\{lt(a,b)|a,b\in \mathcal{C}, a<b\}$ and  $\mathcal{N}=\{lt(a,b)|a,b\in \mathcal{C}, a \ge b\}$. It is easy to verify that the program with rules defined in the following entails all the positive examples and rejects all the negative ones:
\begin{align} 
\text{lessThan}(A,B) &\leftarrow \text{inc}(A,B) \nonumber\\
\text{lessThan}(A,B) &\leftarrow \text{lessThan}(A,C), \text{inc}(C,B) \label{eq:lessThan}
\end{align}
In most ILP systems, the set of possible atoms that can be used in the body of each rule are generated by using a template (e.g. mode declarations in Progol and meta-rules in Metagol (\cite{metagol})). If we allow for $num\_var^i(p)$ variables in the body of the $i^{th}$ rule for the predicate $p$ (e.g., $num\_var^1(lt)=2$, $num\_var^2(lt)=3$ in above example), 
the set of possible (symbolic) atoms for the $i^{th}$ rule for the predicate $p$ is given by:
\begin{align}
&\mathbb{I}^i_{p} = \bigcup_{p^* \in \mathbb{P}} \mathbb{T} ( p^*, V^i_p)  \, \text{, where}\\
&\mathbb{T}(p,V) = \{ p(arg) | \,arg \in Perm(V,\,arity(p)\,)\,\}
\end{align}
where $V^i_p$ is the set of variables ($|V^i_p|=num\_var^i(p))$ for the $i^{th}$ rule and  $\mathbb{P}$  is the set of all the predicates in the program. Further the function $Perm(S,n)$ generates the set of all the permutations of tuples of length $n$ from the elements of a set $S$ and the function $arity(p)$ returns the number of arguments in predicate $p$.
For example, in the \texttt{lessThan} program, $V^1_{lt}=\{A,B\}$ and $V^2_{lt}=\{A,B,C\}$. Consequently, the set of possible atoms can be enumerated as:
\begin{align*}
\mathbb{I}^1_{lt} &= \{\text{inc}(A,A),\text{inc}(A,B),\text{inc}(B,A),\text{inc}(B,B)\} \,\bigcup \,\{lt(A,A),lt(A,B),lt(B,A),lt(B,B)\}    \\
\mathbb{I}^2_{lt} &= \{\text{inc}(A,A),\text{inc}(A,B),\text{inc}(A,C),\dots,\text{inc}(C,C)\} \,\bigcup \,\{lt(A,A),lt(A,B),\dots,\,lt(C,C)\}    
\end{align*}
In general, there are two main approaches to ILP. The bottom-up family of approaches (e.g. Progol) start by examining the provided examples and extract specific clauses from those and try to generalize from those specific clauses. In the top-down approaches (e.g., most neural implementations as well as Metagol and dILP (\cite{evans2018learning})), the possible clauses are generated via a template and the generated clauses are tested against positive and negative examples. Since the space of possible clauses are vast, in most of these systems, very restrictive template rules are employed to reduce the size of the search space. For example, dILP allows for clauses of at most two atoms and only two rules per each predicate. In the above example, since $|\mathbb{I}^2_{lt}|=18$, this corresponds to considering only $\binom{18}{2}$ items from all the possible clauses (i.e., the power set of  $\mathbb{I}^2_{lt}$). Metagol employs a more flexible approach by allowing the programmer to define the rule templates via some meta-rules. However, in practice, this approach does not resolve the issue completely. Even though it allows for more flexibility, defining those templates is itself a complicated task which requires expert knowledge and possible trials and it can still lead to exponentially large space of possible solutions. Later we will consider examples where these kinds of approaches are practically impossible.

Alternatively, we propose a novel approach which allows for learning any arbitrary Boolean function involving several atoms from the set $\mathbb{I}^i_{p}$. This is made possible via a set of differentiable neural functions which can explicitly learn and represent Boolean functions.

\subsection{Differentiable Neural Logic Networks} 
\label{sec:LogicLayer}
Any Boolean functions can be learned (at least in theory) via a typical MLP network. However, since the corresponding logic is stored implicitly in weights of the MLP network, it is very difficult (if not impossible) to decipher the actual learned function. Therefore, MLP is not a good candidate to use in our ILP solver. Our intermediate goal is to design new neuronal functions which are capable of learning and representing Boolean functions in an \emph{explicit} manner. Since any Boolean functions can be expressed in a Disjunctive Normal Form (DNF) or alternatively in a Conjunctive Normal Form (CNF), we first introduce novel conjunctive and disjunctive neurons. We can then combine these elementary functions to form more expressive constructs such as DNF and CNF functions.
We use the extension of the Boolean to real values in the range $[0,1]$ and we use 1 (True) and 0 (False) representations for the two states of a binary variable. We also define the fuzzy unary and dual Boolean functions of two Boolean variables $x$ and $y$ as:
\begin{align}
	\label{eq:BoolAlgebra}
	\bar{x} =  1 - x  \qquad,\quad	x \wedge y =   xy  \qquad 
	x \vee y  = 1 - ( 1 - x )( 1 - y)  
\end{align}
%
This algebraic representation of the Boolean logic allows us to manipulate the logical expressions via Algebra. 
Let $\bx^n \in \{0,1\}^n$ be the input vector for our logical neuron.
\begin{figure}[tb]
	\centering
	\subfloat[][]{
		\small
		\vspace{-5mm}
		\begin{tabular}{|c|c|c|}
			\hline  
			$x_i$ & $m_i$ & $F_c$ \\ 	\toprule  
			0 & 0 & 1 \\ \hline
			0 & 1 & 0 \\ \hline
			1 & 0 & 1 \\ \hline
			1 & 1 & 1 \\ \hline
		\end{tabular}
		\label{fig:Fc}%
	}%
	\qquad
	\subfloat[][]{
		\small
		\begin{tabular}{|c|c|c|}
			\hline	 
			$x_i$ & $m_i$ & $F_d$ \\   	\toprule
			0 & 0 & 0 \\ \hline
			0 & 1 & 0 \\ \hline
			1 & 0 & 0 \\ \hline
			1 & 1 & 1 \\ \hline
		\end{tabular}
		\label{fig:Fd}%
	}
	\caption{Truth table of $F_c(\cdot)$ and $F_d(\cdot)$ functions}%
	\label{fig:FcFd}%
\end{figure}
%
%
In order to implement the conjunction function, we need to select a subset in $\bx^n$ and apply the fuzzy conjunction (i.e. multiplication) to the selected elements. 
To this end, we associate a trainable Boolean membership weight $m_i$ to each input elements $x_i$ from vector $\bx^n$. Further, we define a Boolean function $F_c(x_i,m_i)$ with the truth table as in Fig.\ref{fig:Fc} which is able to include (exclude) each element in (out of) the conjunction function. This design ensures the incorporation of each element $x_i$ in the conjunction function only when the corresponding membership weight is $1$. Consequently, the  neural conjunction function $f_{conj}$ can be defined as:
\vspace{-2mm}
\begin{align}\label{eq:conj}
f_{conj}(\bx^n) &= \prod_{i=1}^{n} F_c(x_i,m_i) \,\,\,  \nonumber\\
\text{where, }  F_c(x_i,m_i) &= \overline{x_i \overline{m_i}} = 1 - m_i ( 1 - x_i) \,,
\end{align}
To ensure the membership weights remain in the range $[0,1]$ we apply a sigmoid function to corresponding trainable weights $w_i$ in the neural network, i.e.,  $m_i = sigmoid( c \,w_i )$ where $c \ge 1$ is a constant. 
Similar to perceptron layers, we can stack $N$ conjunction neurons to create a conjunction layer of size $N$. This layer has the same complexity as a typical perceptron layer without incorporating any bias term. More importantly, this  implementation of the conjunction function makes it possible to interpret the learned Boolean function directly from the values of the membership weights $m_i$. 
The disjunctive neuron can be defined similarly but using the function $F_d$ with truth table as depicted in Fig.\ref{fig:Fd}, i.e.: 
%
\begin{align}\label{eq:disj}
f_{disj}(\bx^n) = \overline { \prod_{i=1}^{n} \overline {F_d(x_i,m_i)} } &=  1 -  \prod_{i=1}^{n} ( 1 - F_d(x_i,m_i) )  \,, \nonumber \\
\text{where, } F_d(x_i,m_i) &= x_i m_i 
\end{align}
%
We call a complex networks made by combining the elementary conjunctive and disjunctive neurons, a dNL (differentiable Neural Logic) network. For example, by cascading a conjunction layer with one disjunctive neuron we can form a dNL-DNF construct. Similarly, a dNL-CNF can be constructed. 

\subsection{ILP as a Satisfiability Problem}
We associate a dNL (conjunction) function ${\mathscr{F}^i_p}$ to $i^{th}$ rule of every intensional predicate $p$ in our logic program. \texttt{intensional} predicates can use other predicates and variables in contrast to the \texttt{extensional} predicates which are entirely defined by the ground atoms. We view the membership weights $m$ in the conjunction neuron as a Boolean flags that indicates whether each atom in a rule is \texttt{off} or \texttt{on}. In this view, the problem of ILP can be seen as finding an assignment to these membership Boolean flags such that the resulting rules applied to the background facts, entail all positive examples and reject all negative examples. However, by allowing these membership weights to be learnable weights, we are formulating a continuous relaxation of the satisfiability problem. This approach is in some ways similar to the approach in dILP \cite{evans2018learning}, but differs in how we define Boolean flags. In dILP, a Boolean flag is assigned to each of the possible combinations of two atoms from the set $\mathbb{I}^i_{p}$. They then use a softmax network to learn the set of winning clauses and they interpret those weights in the softmax network as the Boolean flags that select one clause out of possible clauses. However, as mentioned earlier, in our approach the membership weights of the conjunction (or any other logical function from dNL) can be directly interpreted as the flags in the satisfiability interpretation. 

\subsection{Forward Chaining}
\begin{figure*}
  \includegraphics[width=0.95\textwidth]{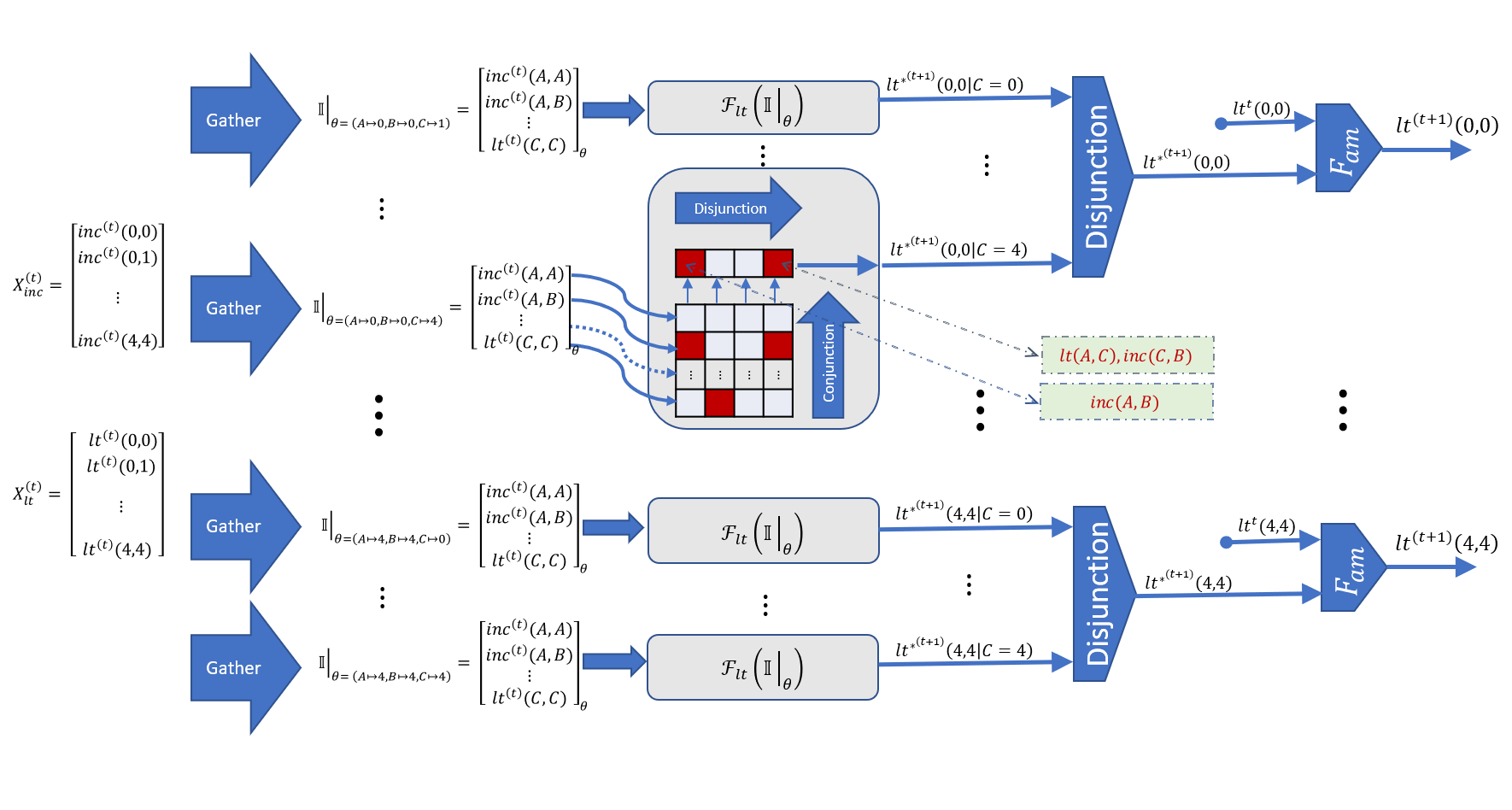}
  \vspace{-5mm}
  \caption{The diagram for one step forward chaining for predicate \texttt{lt} where $\mathscr{F}_{lt}$ is implemented using a dNL-DNF network.}
  \label{fig:ilp_diag}
\end{figure*}

We are now able to formulate the ILP problem as an end-to-end differentiable neural network. 
We associate a (fuzzy) value vector for each predicate $p$ at time-stamp $t$ as $X_p^{(t)}$ which holds the (fuzzy) Boolean values of all the ground atoms involving that predicate.
For the example in consideration (i.e., \texttt{lessThan}), the vector $X^{(t)}_{\text{inc}}$ includes the Boolean values for atoms in $\{\text{inc}(0,0),\text{inc}(0,1),\dots,\text{inc}(4,4)\}$. For extensional predicates, these values will be constant during the forward chain of reasoning, but for intensional predicates such as \texttt{lt}, the values of the $X^{(t)}_{p}$ would change during the application of the predicate rules $\mathscr{F}^i_{p}$ at each time-stamp.
Let $G$ be the set of all ground atoms and $G_p$ be the subset of $G$ associated with predicate $p$. 
For every ground atom $e\in G_p$ and for every rule ${\mathscr{F}^i_p}$, let $\Theta_p^i(e)$ be the set of all the substitutions of the constants into the variables $V_p^i$ which would result in the atom $e$. In the \texttt{lessThan} program (see page 2) for example, for the ground atom $lt(0,2)$, the set of all substitutions corresponding to the second rule (i.e., $i=2$) is given by $\Theta_{lt}^2( \,lt(0,2)\,) = \{ \{A\mapsto 0,B\mapsto 2,C\mapsto 0\}, \dots,\{A\mapsto 0,B\mapsto 2,C\mapsto 4\} \}$. 
We can now define the one step forward inference formula as:
\begin{subequations}
\begin{align}
    &\forall e \in G_p,   X_p^{(t+1)}[e] = F_{am} (X_p^{(t)}[e]  , \mathcal{F}(e ) ) \,\text{, where} \\
    &\mathcal{F}(e) = \bigvee_i \bigvee_{\theta \in \Theta_p^i(e)} \mathscr{F}^i_p (\, \mathbb{I}^i_{p}\rvert_{\theta}\, )\label{eq:inf:2}
\end{align}\label{eq:inference}
\end{subequations}
For the most practical purposes we can assume that the amalgamate function $F_{am}$ is simply the fuzzy disjunction function, but we will consider other options in the Appendix \ref{appx:amalgamate}. Here, for brevity we did not introduce the indexing notations in (\ref{eq:inference}). By $X_p[e]$, we actually mean $X_p[ index(X_p,e)]$ where $index(X_p,e)$ returns the index of the corresponding element of vector $X_p$.  Further, each $\mathscr{F}^i_p$ is the corresponding predicate rule function implemented as a differentiable dNL network (e.g., a conjunctive neuron). In each substitution, this function is applied to the input vector $\mathbb{I}^i_{p}\rvert_{\theta}$ which is evaluated for the substitution $\theta$. As an example, for the ground atom $lt(0,2)$ in the previous example, and for the substitution $\theta =\{A\mapsto 0,B\mapsto 2\}$ corresponding to the first rule we have:
\begin{equation*}
\mathbb{I}^1_{lt} = \{X_{inc}[(0,0)],X_{inc}[(0,2)],X_{inc}[(2,0)],X_{inc}[(2,2)],X_{lt}[(0,0)],X_{lt}[(0,2)],X_{lt}[(2,0)],X_{lt}[(2,2)]\}
\end{equation*}
Fig.\ref{fig:ilp_diag} shows one step forward chaining for learning the predicate $lt$. In this diagram two rules are combined and replaced by one dNL-DNF functions.

\subsection{Training}
We obtain the initial values of the valuation vectors from the background atoms. i.e.,
\begin{equation}
    \forall p, \forall e \in G_p,\quad  \text{ if  } e \in \mathcal{B}, \quad X^{(0)}_p[e] = 1 , \quad   \quad else\quad X^{(0)}_p[e] =0
\end{equation}
We interpret the final values of $X^{(t_{max})}_p[e]$ (after $t_{max}$ steps of forward chaining) as the conditional probability for the value of atom given the model parameters and we define the loss as the average cross-entropy loss between the ground truth provided by the positive and negative examples for the corresponding predicate $p$) and $X_{p}^{(t_{max})}$ which is the algorithm output after $t_{max}$ forward chaining steps.
We train the model using ADAM (\cite{KingmaB14}) optimizer to minimize the aggregate loss over all intensional predicates with the learning rate of 0.001 (in some cases we may increase the rate for faster convergence).
After the training is completed, a zero cross-entropy loss indicates that the model has been able to satisfy all the examples in the positive and negative sets. However, there might exist a few atoms with membership weights of '1' in the corresponding dNL network for a predicate which are not necessary for the satisfiability of the solution. 
However, since there is no gradient at this point, those terms cannot be directly removed during the gradient descent algorithm unless we include some penalty terms.
In practice, we use a simpler approach.
In the final stage of algorithm we remove each atom if by switching its membership variable from ’1’ to ’0’, the loss function does not change.

\subsection{Predicate rules (\texorpdfstring{$\mathscr{F}^i_p$}{Lg})}
    In the majority of the ILP systems, the body of the rules are defined as the conjunction of some atoms. However,in general the predicate rules can be defined as any arbitrary Boolean function of the elements of set $\mathbb{I}_{p}$. One of the main reasons for restricting the form of these rules in most ILP implementations is the vast space of possible Boolean functions that is needed to be considered. For example, by restricting the form of rule's body to a pure Horn clause we reduce the space of possible functions from $2^{2^L}$ to only $2^L$, where $L=| \mathbb{I}^i_{p}|$. Most ILP systems apply much further restrictions. For example, dILP limits the possible combinations to the $\binom{L}{2}$ possible combinations of terms made of two atoms.
    In contrast, in our proposed framework via dNL networks, we are able to learn arbitrary functions with any number of atoms in the formula. Though some functions from the possible $2^{2^L}$ functions require exponentially large number of terms if expressed in DNF form for example, in most of the typical scenarios, a dNL-DNF function with reasonable number of disjunction terms is capable of learning the required logic. 
    Further, even though our approach allows for multiple rules per predicates, in most scenarios we can learn all the rules for a predicate as one DNF formula instead of learning separate rules. 
    Finally, we can easily allow for including the negation of each atom in the formula by concatenating the vector $\mathbb{I}_{p}\rvert_{\theta}$ and its fuzzy negation, i.e., $(1.0-\mathbb{I}_{p}\rvert_{\theta})$ as the input to the $\mathscr{F}^i_p$ function. This would only double the number of parameters of the model. In contrast, in most other implementations of ILP, this would increase the number of parameters and the problem complexity at much higher rates.
\subsection{Implementation and Performance}
We have implemented\footnote{The python implementation of dNL-ILP  
is available at \url{https://github.com/apayani/ILP}} the dNL-ILP solver model using Tensorflow (\cite{tensorflow2015-whitepaper}). In the previous sections, we have outlined the process in a sequential manner. However, in the actual implementations we first create index matrices using all the background facts before starting the optimization task. Further, all the substitution operations for each predicate (at each time-stamp) are carried using a single \texttt{gather} function. Finally, at each time-stamp and for each intensional predicate, all instances of applying (executing) the neural function $\mathscr{F}^i_p$ are carried in a batch operation and in parallel. 
The proposed algorithm allows for a very efficient learning of arbitrary complex formulas and significantly reduces the complexity that arises in the typical ILP systems when increasing the number of possible atoms in each rule. Indeed, in our approach, usually there is no need for any tuning and parameter specification other than the size of the DNF network (total number of rules for a predicate) and specifying the number of  existentially quantified variables for each rule.
On the other hand, since we use a propositionalization step (typical to almost all neural ILP solvers), special care is required when the number of constants in the program is very large. While for the extensional and target predicates we can usually define the vectors $X_p$ corresponding only to the provided atoms in the sets $\mathcal{B}$, $\mathcal{P}$ and $\mathcal{N}$, for the auxiliary predicates we may need to consider many intermediate ground atoms not included in the program. In such cases, when the space of possible atoms is very large, we may need to restrict the set of possible ground atoms.
\section{Past Works}
\label{sec:PASTWORKS}
Addressing all the important past contributions in ILP is a tall order and given the limited space we will only focus on a few recent approaches that are in some ways relevant to our work. Among the ILP solvers that are capable of learning recursive predicates (in an explicit and symbolic manner),  the most notable examples are Metagol (\cite{metagol}) and dILP (\cite{evans2018learning}). Metagol is a powerful method that is capable of learning very complex tasks via using the user-provided meta-rules. The main issue with Metagol is that while it allows for some flexibility in terms of providing the meta-rules, it is not always clear how to define those meta formulas. In practice, unless the expert already has some knowledge regarding the form of the possible solution, it would be very difficult to use this method. dILP, on the other hand, is a neural ILP solvers that, like our method, uses propositionalization of the data and formulates a differentiable neural ILP solver. Our proposed algorithm is in many regards similar to dILP. However, because of the way it define templates, dILP is limited to learning simple predicates with arity of at most two and with maximum two atoms in each rule. CILP++ (\cite{francca2014fast}) is another noticeable neural ILP solver which also uses propositionalization similar to our method and dILP. CLIP++ is a very efficient algorithm and is capable of learning large scale relational datasets. However, since this algorithm uses the bottom clause propositionalization, it is not able to learn recursive predicates. 
In dealing with  uncertain data and specially in the tasks involving classification of the relational datasets, the most notable framework is the probabilistic ILP (PILP) (\cite{de2008probabilistic}) and its variants and also Markov Logic Networks (MLN) \cite{richardson2006markov}. These types of algorithms extend the framework of ILP to handle uncertain data by introducing a probabilistic framework. Our proposed approach is related to PILP in that we also associate a real number to each atom and each rule in the formula. We will compare the performance of our method to this category of statistical relational learners later in our experiment. The methods in this category in general are not capable of learning recursive predicates.

\section{Experiments}
\label{sec:experiments}
The ability to learn recursive predicates is fundamental in learning a variety of algorithmic tasks (\cite{tamaddoni2015towards,cropper2015learning}). In practice, Metagol is the only notable ILP solver which can efficiently learn recursive predicates (via meta-rule templates).
Our evaluations\footnote{Many of the symbolic tasks used in \cite{evans2018learning} as well as some others are provided in the accompanying source code.} show that the proposed dNL-ILP solver can learn a variety of discrete algorithmic tasks involving recursion very efficiently and without the need for predefined meta-rules. Here, we briefly explore two synthetic learning tasks before considering large-scale tasks involving relational datasets.
%
%
%
%
%
%
%
%
%
%
%
%
\label{subsec:Benchmarks}

\subsection{Learning decimal multiplication}
%
%
%
%
We use dNL-ILP solver for learning the predicates $mul/3$ for decimal multiplication using only the positive and negative examples.
We use $\mathcal{C}=\{0,1,2,3,4,5,6\}$ as constants and our background knowledge is consisted of the extensional predicates $\{zero/1,inc/2,add/3\}$, where  $inc/2$ defines increment of one and $add/3$ defines the addition. The target predicate is $mul(A,B,C)$ and we allow for using 5 variables (i.e., $num\_var^i(mul)=5$) in each rule. We use a dNL-DNF network with 4 disjunction terms (4 conjunctive rules) for learning $\mathscr{F}_{mul}$. It is worth noting that since we do not know in advance how many rules would be needed, we should pick an arbitrary number and increase in case the ILP program cannot explain all the examples. Further, we set the $t_{max}=8$.
One of the solutions that our model finds is:
\begin{align*}
\vspace{-3mm}
mul(A,B,C) &\leftarrow zero(B), \, zero(C) \\
mul(A,B,C) &\leftarrow mul(A,D,E),  inc(D,B), add(E,A,C)
\end{align*}
%
%
%
%
%
\subsection{Sorting}
\label{subsec:sorting}
The sorting task is more complex than the previous task since it requires not only the list semantics, but also many more constants compared to the arithmetic problem. We implement the list semantic by allowing the use of functions in defining predicates. For a data of type \texttt{list}, we define two functions $H$ and $t$ which allow for decomposing a list into head and tail elements, i.e $A=[A_H|A_t]$.
We use elements of $\{a,b,c,d\}$ and all the ordered lists made from permutations of up to three elements as constants in the program (i.e., $|\mathcal{C}|=40)$. We use extensional predicates such as $gt$ (greater than), $eq$ (equals) and $lte$ (less than or equal) to define ordering between the elements of lists as part of the background knowledge. We allow for using 4 variables (and their functions) in defining the predicate $sort$ (i.e., $num\_var^i(sort)=4$). One of the solution that our model finds is:
\topskip=2pt
\begin{align*}
sort(A,B) &\leftarrow sort(A_H,C),\, lte(C_t,A_t), eq(B_H,C),\,eq(A_t,B_t) \\
sort(A,B) &\leftarrow sort(A_H,C),\, sort(D,B_H),\, gt(C_t,A_t),\, eq(B_t,C_t),\,  eq(D_H,C_H),\, eq(A_t,D_t)
\end{align*}

Even though the above examples involve learning tasks that may not seem very difficult on the surface, and deal with relatively small number of constants, they are far from trivial.
To the best of our knowledge, learning a recursive predicate for a complex algorithmic task such as \texttt{sort} which involves multiple recursive rules with 6 atoms and includes $12$ variables (by counting two functions head and tail per variables) is beyond the power of any existing ILP solver. Here for example, the total number of possible atoms to choose from is $\lvert\mathbb{I}^2_{sort}\rvert = 176$ and for the case of choosing 6 elements from this list we need to consider $\binom{176}{6} > 3\times 10^{10}$ possible combinations (assuming we knew in advance there is a need for 6 atoms). While we can somewhat reduce this large space by removing some of the improbable clauses, no practical ILP solver is capable of learning these kinds of relations directly from examples.

\subsection{Classification for Relational Data}
\label{subsec:relational}
We evaluate the performance of our proposed ILP solver in some benchmark ILP tasks. We use relational datasets Mutagenesis (\cite{Mutagenesis}), UW-CSE (\cite{richardson2006markov}) as well as IMDB and Cora datasets\footnote{Publicly-available at \url{https://relational.fit.cvut.cz/}}. Table \ref{tbl:datasets} summarizes the features of these datasets.
\begin{table}[ht]
	\caption{Dataset Features}
    \label{tbl:datasets}
	\centering 
	\begin{tabular} { l  l l l l l}
		\toprule
		Dataset &  Constants & Predicates & Examples & Target Predicate \\ 		
		\midrule
		Mutagenesis & 7045 & 20 & 188 & $active(A)$ \\
		UW-CSE & 7045 & 15 & 16714 & $advisedBy(A,B)$ \\
		Cora & 3079 & 10 & 70367 & $sameBib(A,B)$ \\
		IMDB & 316 & 10 & 14505 & $workingUnder(A,B)$ \\
		\bottomrule
	\end{tabular}
\end{table}
As baseline we are comparing our method with the state of the art algorithms based on Markov Logic Networks such GSLP (\cite{dinh2011generative}), LSM (\cite{kok2009learning}), MLN-B (Boosted MLN), B-RLR (\cite{ramanan2018structure}) as well as  probabilistic ILP based algorithms such as SleepCover (\cite{bellodi2015structure}). 
Further, since in most of these datasets, the number of negative examples are significantly greater than the positive examples, we report the Area Under Precision Recall (AUPR) curve as a more reliable measure of the classification performance. We use 5-fold cross validations except for the Mutagenesis dataset which we have used 10-fold and we report the average AUPR over all the folds. Table \ref{tbl:results_classification} summarizes the classification performance for the 4 relational datasets. 
\begin{table}[ht]
	\caption{AUPR measure for the 4 relational classification tasks}
    \label{tbl:results_classification}
	\centering 
	\begin{tabular} { l  c  c c c c c c }
		\toprule
		Dataset &  GSLP & LSM & SleepCover  & MLN-B & B-RLR & dNL-ILP \\ 		
	    \midrule
		Mutagenesis & 071  & 0.76 & 0.95 &N/A&N/A           & \textbf{0.97} \\
		UW-CSE      & 0.42 & 0.46 & 0.07 &\textbf{0.91} & 0.89 & 0.51\\
		Cora        & 0.80 & 0.89 &  N/A & N/A  & N/A & \textbf{0.95}\\
		IMDB        & 0.71 & 0.79 & N/A  & 0.83 & 0.90 & \textbf{1.00}\\
		\bottomrule
	\end{tabular}
\end{table}
As the results show, our proposed method outperforms the previous algorithms in the three tasks; Mutagenesis, Cora and IMDB. In case of IMDB dataset, it reaches the perfect classification (AUROC=1.0, AUPR=1.0). This impressive performance is only made possible because of the ability of learning recursive predicates. Indeed, when we disallow the recursion in this model, the AUPR performance drops to $0.76$.
The end-to-end design of our differentiable ILP solver makes it possible to combine some other forms of learnable functions with the dNL networks. For example, while handling continuous data is usually difficult in most ILP solvers, we can directly learn some threshold values to create binary predicates from the continuous data (see Appendix \ref{appx:continuous})\footnote{Alternatively, we can assign learnable probabilistic functions to those variables (see Appendix \ref{appx:uncertain}).}. We have used this method in the Mutagenesis task to handle the continuous data in this dataset. For the case of UW-CSE, however, our method did not perform as well. 
One of the reasons is arguably the fact that the number of negative examples is significantly larger than the positive ones for this dataset.
Indeed, in some of the published reports, (e.g. \cite{francca2014fast}), the number of negative examples are limited using the closed world assumption as \cite{davis2005integrated}. Because of the difference  in hardware, it is difficult to directly compare the speed of algorithms. In our case, we have evaluated the models using a 3.70GHz CPU, 16GB RAM and GeForce GTX 1080TI graphic card. Using this setup the problems such as IMDB, Mutagenesis are learned in just a few seconds. For Cora, the model creation takes about one minute and the whole simulation for any fold takes less than 3 minutes.
\section{Conclusion}
\label{sec:conclusion}
We have introduced dNL-ILP as a new framework for learning inductive logic programming problems. Using various experiments we showed that dNL-ILP outperforms past algorithms for learning algorithmic and recursive predicates. Further, we demonstrated that dNL-ILP is capable of learning from uncertain and relational data and outperforms the state of the art ILP solvers in classification tasks for Mutagenesis, Cora and IMDB datasets.
 
\newpage

\appendix
\label{appx:notations}
\section{Notations}

\begin{table}[ht]
	\caption{Some of the notations used in this paper}
    \label{tbl:notations}
	\centering 
	\begin{tabular} { l  l }
		\toprule
		Notation &  Explanation \\ 		
			\midrule
		$p/n$   & a predicate $p$ of arity $n$\\  
 		$\mathcal{C}$ &    the set of constants in the program \\
		$\mathcal{B}$ &    the set of background atoms \\
		$\mathcal{P}$ &    the set of positive examples \\
		$\mathcal{N}$ &    the set of negative examples \\
		$G_p$ &    the set of ground atoms for predicate $p$ \\
		$G$ &    the set of all ground atoms\\
		 $\mathbb{P}$ & the set of all predicates in the program \\
		$X_p^{(t)}$  & the fuzzy values of all the ground atoms for predicate $p$ at time $t$ \\
		$A \mapsto a$  & substitution of constant $a$ into variable  $A$\\
		$\mathcal{R}$  & the set of all the rules in the program\\
		$perm(S,n)$ & the set of all the permutations of tuples of length $n$ from the set $S$ \\
		$\mathbb{T}(p,V)$ & the set of atoms involving predicate $p$ and using the variables in the set $V$ \\
		$\mathbb{I}^i_{p}$ & the set of all atoms that can be used in generating $i^{th}$ rule for predicate $p$ \\
		\bottomrule
	\end{tabular}
\end{table}

\section{Amalgamate Function  (\texorpdfstring{$F_{am}$}{Lg})} 
In most scenarios we may set this function to a disjunction functions. However, we can modify this function for some specific purposes. For example:
\begin{itemize}
    \item $F_{am}(old,new)=old \bigwedge new$: by this choice we can implement a notion of $\forall$(for all) in logic which can be useful in certain programs. There is an example in the source code which learns array indexing by the help of this option.
    \item $F_{am}(old,new)=new$: by this choice we can learn transient logic (an alternative approach to the algorithm presented in \cite{inoue2014learning}).
\end{itemize}
\label{appx:amalgamate}


\section{Dream4 challenge Experiment (handling uncertain data)}
\label{appx:uncertain}

Inferring the causal relationship among different genes is one of the important problems in biology. In this experiment, we study the application of dNL-ILP for inferring the structure of gene regulatory networks using 10-genes time-series dataset from the DREAM4 challenge tasks \cite{marbach2009dream4}.
In the 10-gene challenge, the data consists of 5 different biological systems, each composed of 10 genes. For each system a time-series containing 105 noisy readings of the genes expressions (in range $[0,1]$) is provided. The time-series is obtained via 5 different experiments, each created using a simulated perturbation in a subset of genes and recording the gene expressions over time.
To tackle this problem using dNL-ILP framework, we simply assume that each gene can be in one of the two states: \texttt{on} (excited or perturbed) or \texttt{off}. The key idea here is to model each gene's state (\texttt{off} or \texttt{On}) using two different approaches and then aim to get a consensus on the gene's state using these two different probes.
To accomplish this, for each gene $G_i$ we define the predicate ${\text{off}}_i$ which evaluates the state of $G_i$ using the corresponding continuous values. We also use predicate ${\text{inf\_off}}_i$ which takes the state of all the predicates $\text{off}_j$'s ($j \neq i$) to infer the state of $G_i$.
To ensure that at each background data ${\text{inf\_off}}_i$ would be close to $\text{off}_i$, we define another auxiliary predicate $aux_i$ with predicate function defined as $\mathscr{F}_{{aux}_i}= 1 - \lvert {\text{inf\_off}}_i - \text{off}_i\rvert$.

Since the state of genes are uncertain, we use a probabilistic approach and assume that each gene state is conditionally distributed according to a Gaussian mixture (with 4 components), i.e.,  $x|\text{off} \sim \text{GMM}_{\text{off}}$ and $x|\text{on} \sim \text{GMM}_{\text{on}}$. As such, we design $\mathscr{F}_{{\text{inf\_off}}_i}$ such that it returns the probability of the $G_i$ gene being \texttt{off}, and  let all the parameters of the mixture models be trainable weights.
For the $\mathscr{F}_{{\text{Inf\_off}}_i}$ we use a dNL-CNF network with only one term.
Each data point in the time series corresponds to one background knowledge and consists of the continuous value of each gene expression. For each gene, we assign 5 percent of data points with the lowest and highest absolute distance from mean as positive and negative examples for predicate $\text{off}_i$, respectively. We interpret the values of the membership weights in the trained dNL-CNF networks which are used in $\mathscr{F}_{{\text{Inf-off}}_i}$ as the degree of connection between two genes. Table \ref{tbl:gene} compares the performance of dNL-ILP to the two state of the art algorithms  NARROMI \cite{zhang2012narromi} and MICRAT \cite{yang2018micrat} for  10-gene classification tasks of DREAM4 dataset.
	\begin{table}[h]
	\caption{DREAM4 challenge scores}
 
    \label{tbl:gene}
	\centering
	\small
	\begin{tabular} {|c| c |c|c|}
		\hline
		\diagbox{Metric}{Method}   & NARROMI & MICRAT & dNL-ILP \\	\hline
		Accuracy & 0.82  & \textbf{0.87} & 0.86\\ 
		\hline
		F-Score & 0.35  & 0.32 & \textbf{0.36}\\
		\hline
		MCC & 0.24  & 0.33 & \textbf{0.35}\\ 
        \hline
	\end{tabular}
\end{table}
\section{UCI Dataset Classification (handling continuous data)}
\label{appx:continuous}

%
%
%
%
%
%
Reasoning using continuous data has been an ongoing challenge for ILP. Most of the current approaches either model continuous data as random variables and use probabilistic ILP framework \cite{de2008probabilistic}, or use some forms of discretization using iterative approaches. The former approach cannot be applied to the cases where we do not have reasonable assumptions for the probability distributions. Further, the latter approach is usually limited to small scale problems (e.g. \cite{ribeiro2017inductive}) since the search space grows exponentially as the number of continuous variables and the boundary decisions increases. Alternatively, the end-to-end design of dNL-ILP makes it rather easy to handle continuous data. Recall that even though we usually use dNL based functions, the predicate functions $\mathscr{F_p}'s$ can be defined as any arbitrary Boolean function in our model. Thus, for each continuous variable $x$ we define $k$ lower-boundary predicates $gt_{x_i}(x,l_{x_i})$ as well as $k$ upper-boundary predicates $lt_{x_i}(x,u_{x_i})$ where $i\in\{1,\dots,k\}$. We let the boundary values $l_{x_i}$'s and $u_{x_i}$'s be trainable weights and we define the upper-boundary and lower-boundary predicate functions as:
\begin{equation*}
    \mathscr{F}_{{gt_x}_i} = \sigma ( c \, (x-u_{x_i})) \,\,,\,\,
    \mathscr{F}_{{lt_x}_i} = \sigma ( - c \, (x-l_{x_i})),
\end{equation*}
where $\sigma$ is the sigmoid function and $c \gg 1$ is a constant.   
To evaluate this approach we use it in a classification task for two datasets containing continuous data; Wine and Sonar from UCI Machine learning dataset \cite{Dua:2017} and compare its performance to the ALEPH \cite{srinivasan2001aleph}, a state-of-the-art ILP system, as well as the recently proposed FOLD+LIME algorithm \cite{shakerin2018induction}. 
The wine classification task involves 13 continuous features and  three classes and the Sonar task is a binary classification task involving 60 features. For each class we define a corresponding intensional predicate via dNL-DNF and learn that predicate from a set of $lt_{x_i}$ and $gt_{x_i}$ predicates corresponding to each continuous feature. We set the number of boundaries to 6 (i.e., $k=6$).
The classification accuracy results for the 5-fold cross validation setting is depicted in Table \ref{tbl:continuous}.
	\begin{table}[h]
	\caption{Classification accuracy}
 
    \label{tbl:continuous}
	\centering 
	\small
	\begin{tabular} {|c| c|c |c|}
		\hline
		Task &  ALEPH+LIME & FOLD+LIME & dNL-ILP \\		
		\hline
		Wine &0.92 & 0.93  & \textbf{0.98} \\ 
		\hline
		Sonar& 0.74 &0.78  & \textbf{0.85}  \\   
		\hline
	\end{tabular}
\end{table}
\bibliographystyle{named}
\bibliography{refs_payani}

\begin{thebibliography}{}

\bibitem[\protect\citeauthoryear{Abadi \bgroup \em et al.\egroup
  }{2016}]{tensorflow2015-whitepaper}
Mart{\'\i}n Abadi, Paul Barham, Jianmin Chen, Zhifeng Chen, Andy Davis, Jeffrey
  Dean, Matthieu Devin, Sanjay Ghemawat, Geoffrey Irving, Michael Isard, et~al.
\newblock Tensorflow: A system for large-scale machine learning.
\newblock In {\em 12th $\{$USENIX$\}$ Symposium on Operating Systems Design and
  Implementation ($\{$OSDI$\}$ 16)}, pages 265--283, 2016.

\bibitem[\protect\citeauthoryear{Bader \bgroup \em et al.\egroup
  }{2008}]{bader2008connectionist}
Sebastian Bader, Pascal Hitzler, and Steffen H{\"o}lldobler.
\newblock Connectionist model generation: A first-order approach.
\newblock {\em Neurocomputing}, 71(13-15):2420--2432, 2008.

\bibitem[\protect\citeauthoryear{Bellodi and
  Riguzzi}{2015}]{bellodi2015structure}
Elena Bellodi and Fabrizio Riguzzi.
\newblock Structure learning of probabilistic logic programs by searching the
  clause space.
\newblock {\em Theory and Practice of Logic Programming}, 15(2):169--212, 2015.

\bibitem[\protect\citeauthoryear{Cropper and
  Muggleton}{2015}]{cropper2015learning}
Andrew Cropper and Stephen~H Muggleton.
\newblock Learning efficient logical robot strategies involving composable
  objects.
\newblock In {\em Twenty-Fourth International Joint Conference on Artificial
  Intelligence}, 2015.

\bibitem[\protect\citeauthoryear{Cropper and Muggleton}{2016}]{metagol}
Andrew Cropper and Stephen~H. Muggleton.
\newblock Metagol system.
\newblock https://github.com/metagol/metagol, 2016.

\bibitem[\protect\citeauthoryear{Davis \bgroup \em et al.\egroup
  }{2005}]{davis2005integrated}
Jesse Davis, Elizabeth Burnside, In{\^e}s de~Castro~Dutra, David Page, and
  V{\'\i}tor~Santos Costa.
\newblock An integrated approach to learning bayesian networks of rules.
\newblock In {\em European Conference on Machine Learning}, pages 84--95.
  Springer, 2005.

\bibitem[\protect\citeauthoryear{De~Raedt and
  Kersting}{2008}]{de2008probabilistic}
Luc De~Raedt and Kristian Kersting.
\newblock Probabilistic inductive logic programming.
\newblock In {\em Probabilistic Inductive Logic Programming}, pages 1--27.
  Springer, 2008.

\bibitem[\protect\citeauthoryear{Debnath \bgroup \em et al.\egroup
  }{1991}]{Mutagenesis}
A.~K. Debnath, R.~L. {Lopez de Compadre}, G.~Debnath, A.~J. Shusterman, and
  C.~Hansch.
\newblock {Structure-activity relationship of mutagenic aromatic and
  heteroaromatic nitro compounds. Correlation with molecular orbital energies
  and hydrophobicity.}
\newblock {\em Journal of medicinal chemistry}, 34(2):786--797, 1991.

\bibitem[\protect\citeauthoryear{Dinh \bgroup \em et al.\egroup
  }{2011}]{dinh2011generative}
Quang-Thang Dinh, Matthieu Exbrayat, and Christel Vrain.
\newblock Generative structure learning for markov logic networks based on
  graph of predicates.
\newblock In {\em Twenty-Second International Joint Conference on Artificial
  Intelligence}, 2011.

\bibitem[\protect\citeauthoryear{Dua and Karra~Taniskidou}{2017}]{Dua:2017}
Dheeru Dua and Efi Karra~Taniskidou.
\newblock {UCI} machine learning repository, 2017.

\bibitem[\protect\citeauthoryear{Evans and
  Grefenstette}{2018}]{evans2018learning}
Richard Evans and Edward Grefenstette.
\newblock Learning explanatory rules from noisy data.
\newblock {\em Journal of Artificial Intelligence Research}, 61:1--64, 2018.

\bibitem[\protect\citeauthoryear{Fran{\c{c}}a \bgroup \em et al.\egroup
  }{2014}]{francca2014fast}
Manoel~VM Fran{\c{c}}a, Gerson Zaverucha, and Artur S~d’Avila Garcez.
\newblock Fast relational learning using bottom clause propositionalization
  with artificial neural networks.
\newblock {\em Machine learning}, 94(1):81--104, 2014.

\bibitem[\protect\citeauthoryear{H{\"o}lldobler \bgroup \em et al.\egroup
  }{1999}]{holldobler1999approximating}
Steffen H{\"o}lldobler, Yvonne Kalinke, and Hans-Peter St{\"o}rr.
\newblock Approximating the semantics of logic programs by recurrent neural
  networks.
\newblock {\em Applied Intelligence}, 11(1):45--58, 1999.

\bibitem[\protect\citeauthoryear{Inoue \bgroup \em et al.\egroup
  }{2014}]{inoue2014learning}
Katsumi Inoue, Tony Ribeiro, and Chiaki Sakama.
\newblock Learning from interpretation transition.
\newblock {\em Machine Learning}, 94(1):51--79, 2014.

\bibitem[\protect\citeauthoryear{Kingma and Ba}{2014}]{KingmaB14}
Diederik~P. Kingma and Jimmy Ba.
\newblock Adam: A method for stochastic optimization.
\newblock {\em CoRR}, abs/1412.6980, 2014.

\bibitem[\protect\citeauthoryear{Kok and Domingos}{2009}]{kok2009learning}
Stanley Kok and Pedro Domingos.
\newblock Learning markov logic network structure via hypergraph lifting.
\newblock In {\em Proceedings of the 26th annual international conference on
  machine learning}, pages 505--512. ACM, 2009.

\bibitem[\protect\citeauthoryear{Marbach \bgroup \em et al.\egroup
  }{2009}]{marbach2009dream4}
Daniel Marbach, Thomas Schaffter, Dario Floreano, Robert~J Prill, and Gustavo
  Stolovitzky.
\newblock The dream4 in-silico network challenge.
\newblock {\em Draft, version 0.3}, 2009.

\bibitem[\protect\citeauthoryear{Muggleton}{1995}]{muggleton1995inverse}
Stephen Muggleton.
\newblock Inverse entailment and progol.
\newblock {\em New generation computing}, 13(3-4):245--286, 1995.

\bibitem[\protect\citeauthoryear{Ramanan \bgroup \em et al.\egroup
  }{2018}]{ramanan2018structure}
Nandini Ramanan, Gautam Kunapuli, Tushar Khot, Bahare Fatemi, Seyed~Mehran
  Kazemi, David Poole, Kristian Kersting, and Sriraam Natarajan.
\newblock Structure learning for relational logistic regression: An ensemble
  approach.
\newblock In {\em Sixteenth International Conference on Principles of Knowledge
  Representation and Reasoning}, 2018.

\bibitem[\protect\citeauthoryear{Ribeiro \bgroup \em et al.\egroup
  }{2017}]{ribeiro2017inductive}
Tony Ribeiro, Sophie Tourret, Maxime Folschette, Morgan Magnin, Domenico
  Borzacchiello, Francisco Chinesta, Olivier Roux, and Katsumi Inoue.
\newblock Inductive learning from state transitions over continuous domains.
\newblock In {\em International Conference on Inductive Logic Programming},
  pages 124--139. Springer, 2017.

\bibitem[\protect\citeauthoryear{Richardson and
  Domingos}{2006}]{richardson2006markov}
Matthew Richardson and Pedro Domingos.
\newblock Markov logic networks.
\newblock {\em Machine learning}, 62(1-2):107--136, 2006.

\bibitem[\protect\citeauthoryear{Serafini and Garcez}{2016}]{serafini2016logic}
Luciano Serafini and Artur~d'Avila Garcez.
\newblock Logic tensor networks: Deep learning and logical reasoning from data
  and knowledge.
\newblock {\em arXiv preprint arXiv:1606.04422}, 2016.

\bibitem[\protect\citeauthoryear{Shakerin and
  Gupta}{2018}]{shakerin2018induction}
Farhad Shakerin and Gopal Gupta.
\newblock Induction of non-monotonic logic programs to explain boosted tree
  models using lime.
\newblock {\em arXiv preprint arXiv:1808.00629}, 2018.

\bibitem[\protect\citeauthoryear{Srinivasan}{2001}]{srinivasan2001aleph}
Ashwin Srinivasan.
\newblock The aleph manual, 2001.

\bibitem[\protect\citeauthoryear{Tamaddoni-Nezhad \bgroup \em et al.\egroup
  }{2015}]{tamaddoni2015towards}
Alireza Tamaddoni-Nezhad, David Bohan, Alan Raybould, and Stephen Muggleton.
\newblock Towards machine learning of predictive models from ecological data.
\newblock In {\em Inductive Logic Programming}, pages 154--167. Springer, 2015.

\bibitem[\protect\citeauthoryear{Yang \bgroup \em et al.\egroup
  }{2018}]{yang2018micrat}
Bei Yang, Yaohui Xu, Andrew Maxwell, Wonryull Koh, Ping Gong, and Chaoyang
  Zhang.
\newblock Micrat: a novel algorithm for inferring gene regulatory networks
  using time series gene expression data.
\newblock {\em BMC systems biology}, 12(7):115, 2018.

\bibitem[\protect\citeauthoryear{Zhang \bgroup \em et al.\egroup
  }{2012}]{zhang2012narromi}
Xiujun Zhang, Keqin Liu, Zhi-Ping Liu, B{\'e}atrice Duval, Jean-Michel Richer,
  Xing-Ming Zhao, Jin-Kao Hao, and Luonan Chen.
\newblock Narromi: a noise and redundancy reduction technique improves accuracy
  of gene regulatory network inference.
\newblock {\em Bioinformatics}, 29(1):106--113, 2012.

\end{thebibliography}

\end{document}